\documentclass{article}

\usepackage[preprint]{neurips_2026}

\usepackage[utf8]{inputenc}
\usepackage[T1]{fontenc}
\usepackage{hyperref}
\usepackage{url}
\usepackage{booktabs}
\usepackage{amsfonts}
\usepackage{amsmath}
\usepackage{amssymb}
\usepackage{nicefrac}
\usepackage{microtype}
\usepackage{xcolor}
\usepackage{graphicx}
\usepackage{float}
\usepackage{placeins}
\usepackage[most]{tcolorbox}

\definecolor{promptTint}{HTML}{B77B58}

\newtcolorbox{promptbox}{%
  enhanced, breakable,
  arc=5pt, boxrule=0.7pt,
  colback=promptTint!22!white, colframe=promptTint,
  before skip=8pt, after skip=10pt,
  left=14pt, right=14pt, top=12pt, bottom=12pt,
  parbox=false,
}

\newcounter{promptcounter}
\renewcommand{\thepromptcounter}{\arabic{promptcounter}}
\newcommand{\promptcaption}[1]{%
  \refstepcounter{promptcounter}%
  \par\vspace{4pt}%
  {\small Prompt~\thepromptcounter: #1\par}%
  \vspace{10pt}%
}

\title{DPBench: Structural Determinants of Multi-Agent\\LLM Coordination Under Simultaneous Resource Contention}

\author{
Najmul Hasan \quad Prashanth BusiReddyGari\thanks{
Corresponding author: \texttt{prashanth.busireddygari@uncp.edu}.\\ 
Code is available at: \url{https://github.com/najmulhasan-code/dpbench}
} \\
Department of Mathematics and Computer Science \\
University of North Carolina at Pembroke \\
Pembroke, NC, USA
}

\begin{document}

\maketitle

\begin{figure}[H]
  \centering
  \includegraphics[width=\linewidth]{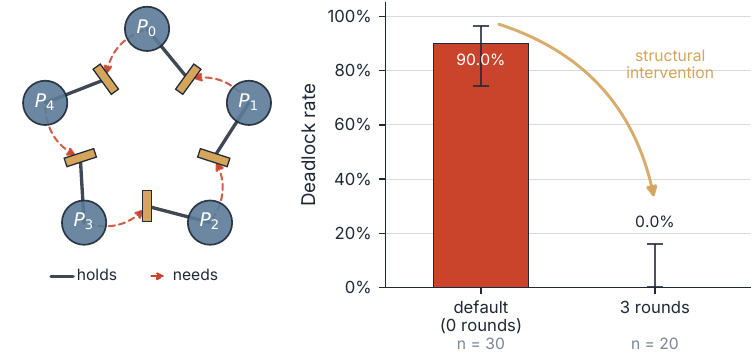}
  \caption{The model is unchanged; the protocol determines the outcome. Left: the $N{=}5$ Dining Philosophers configuration. Right: three rounds of pre-commitment communication drive Gemini~2.5~Flash from $90.0\%$ deadlock to $0.0\%$. Bars are $95\%$ Wilson CIs.}
  \label{fig:teaser}
\end{figure}

\begin{abstract}
We present DPBench, a benchmark for evaluating coordination in multi-agent systems built from large language models. Existing benchmarks measure task-level success under a fixed protocol; the structural conditions under which coordination succeeds or fails at all have not been characterised. DPBench adapts the Dining Philosophers problem into a controlled testbed where the action protocol, the communication structure, and the group size each vary independently. We evaluate six agents: GPT-5.2, Claude Opus 4.5, Grok 4.1, Gemini~2.5~Flash, Llama~4~Maverick, and a uniform-random baseline. Under simultaneous action at $N{=}5$ with the default prompt, deadlock ranges from $25.0\%$ ($95\%$ Wilson CI $[11.2, 46.9]$) for GPT-5.2 to $90.0\%$ $[74.4, 96.5]$ for Gemini~2.5~Flash; sequential action is solved by four of the six. Holding the model fixed at Gemini~2.5~Flash, three protocol variables drive deadlock from $90\%$ to within CI of zero: three rounds of pre-commitment communication ($0.0\%$ vs.\ single-round $86.7\%$), a prompt encoding a classical concurrency primitive ($0.0\%$ for resource-ordering and symmetry-breaking, against $100\%$ for the minimal prompt), or doubling the group from $N{=}5$ to $N{=}10$ ($90.0\%$ to $10.0\%$). Single-round messaging and memory of past timesteps do not change the rate at the sample size we ran. Whether the same model coordinates or deadlocks is determined by the protocol, not by the model's capability.
\end{abstract}

\section{Introduction}

A growing class of LLM systems coordinates several agents rather than answering as a single chat partner. Software-engineering agents hold separate roles around a shared codebase~\citep{hong2024metagpt, agashe2025llm}; retrieval and tool-use pipelines delegate sub-queries between specialised LLM calls~\citep{yao2023react, zhou2024lats}; orchestration frameworks let multiple LLM instances act on a shared environment~\citep{liu2024agentbench, wu2025astra}. Each step in this direction takes the system further from single-turn answering and closer to the kind of distributed-computing problem that has been studied for sixty years. Agents now operate concurrently, share state, and must coordinate to avoid interfering with each other.

The benchmark literature for multi-agent LLM systems mostly measures whether agents \emph{succeed} when given a particular protocol~\citep{liu2024agentbench, hong2024metagpt, zhu2025multiagentbench, duan2024gtbench, du2024debate, wang2024zsceval}. We study a complementary question: under what protocols does coordination succeed at all? Our concern is the kind of failure that distributed-systems engineers call deadlock, the failure that appears when several agents claim shared resources at once and end up in a circular wait that does not resolve. Deadlock is well understood in concurrent computing. Dijkstra's Dining Philosophers~\citep{dijkstra1965solution} formalised the simplest setting where the failure shows up. The classical literature also gave two simple sufficient conditions for deadlock freedom: resource ordering, where every agent picks up resources in a fixed global order, and symmetry breaking, where one agent reverses its grab order. These conditions do not require fairness, scheduling tricks, or even communication; they are statements about the protocol.

\paragraph{The puzzle.} On the same task, the same Gemini 2.5 Flash model deadlocks $90.0\%$ of the time under the default protocol and $0.0\%$ of the time once we add three rounds of pre-commitment messaging. The same model deadlocks $0.0\%$ once we add a one-paragraph prompt encoding the resource-ordering rule. The same model deadlocks $10.0\%$ instead of $90.0\%$ once we change $N$ from $5$ to $10$. The decision-making module is identical in every case; the protocol around it is not. The interesting object is the protocol.

\paragraph{What this paper claims.} Three protocol-level variables determine whether multi-agent LLM coordination succeeds under simultaneous resource contention: pre-commitment communication rounds, the prompt-level coordination strategy, and the group size. Within ranges where these variables permit coordination, every model we test succeeds. Outside those ranges, the same model fails at rates that depend on the model. The contribution is the characterisation of these conditions, not the claim that LLMs cannot coordinate.

\paragraph{What the paper shows.} In simultaneous mode at $N{=}5$ with the default prompt and no inter-agent communication, deadlock rates range from $25.0\%$ (GPT-5.2) to $90.0\%$ (Gemini~2.5~Flash) across five frontier LLMs. The lowest of these, GPT-5.2 at $25.0\%$ $[11.2, 46.9]$, overlaps the random baseline $13.3\%$ $[5.3, 29.7]$ at $n{=}20\text{--}30$ episodes. Models do not separate cleanly from chance under default conditions. Sequential coordination, where one philosopher acts at a time, is solved by four of six agents (deadlock $0.0\%$ point estimate, $95\%$ Wilson upper CI at or below $16.1\%$). Two anomalies remain: Claude~Opus~4.5 deadlocks $60.0\%$ $[38.7, 78.1]$ and Grok~4.1 $25.0\%$ $[11.2, 46.9]$. Action distributions show \textsc{Wait}-as-default behaviour rather than canonical circular wait; we report both.

Three structural variables drive the simultaneous-coordination failure to within CI of zero for Gemini~2.5~Flash, the agent that fails most. Three rounds of pre-commitment discussion drive deadlock from $90.0\%$ to $0.0\%$. Prompts that encode resource ordering or symmetry breaking each reach $0.0\%$. Increasing $N$ from $5$ to $10$ takes deadlock from $90.0\%$ to $10.0\%$. By contrast, single-round messaging, one batch of messages exchanged just before agents commit, does \emph{not} statistically reduce deadlock at $n{=}30$ ($86.7\%$ $[70.3, 94.7]$ vs.\ baseline $90.0\%$ $[74.4, 96.5]$). The structure of the communication matters; one round is not enough. Memory of past timesteps without multi-round communication produces no detectable effect either (Appendix~\ref{sec:appendix-memory}).

\section{Coordination as a Structural Problem}
\label{sec:context}

\paragraph{Multi-agent LLM benchmarks.} The current generation of multi-agent LLM benchmarks measures task-level success. AgentBench~\citep{liu2024agentbench} evaluates LLMs as agents across eight environments and reports gaps between closed and open-weight models. MetaGPT~\citep{hong2024metagpt} encodes role-based standardised operating procedures into a software-engineering pipeline. MultiAgentBench~\citep{zhu2025multiagentbench} measures cooperative and competitive behaviour over six environments. GTBench~\citep{duan2024gtbench} evaluates LLMs in game-theoretic interactions. Multi-agent debate~\citep{du2024debate} improves factuality and reasoning by letting several LLM instances argue. ZSC-Eval~\citep{wang2024zsceval} measures zero-shot coordination performance. Specialised settings include medical~\citep{kim2024mdagents}, embodied AI~\citep{yang2024eai}, and assistive collaboration~\citep{hua2024assistive}. Adversarial behaviours have also been reported, including secret collusion among LLM agents~\citep{motwani2024collusion} and behavioural shifts under harm-eliciting prompts~\citep{andriushchenko2025agentharm}. The DPBench question is orthogonal: not what the agents accomplish under a specific protocol, but what aspects of the protocol make coordination feasible at all.

\paragraph{LLM agents, planning, and reasoning brittleness.} A separate line of work has documented brittleness in LLM planning and symbolic reasoning. \citet{valmeekam2023planning} and PlanBench~\citep{valmeekam2024planbench} show that LLMs struggle on classical planning. \citet{kambhampati2024position} argues that LLMs do not plan in a strict sense, and \citet{stechly2025selfverification} shows they cannot reliably self-verify. \citet{mirzadeh2024gsmsymbolic} demonstrates the same model collapsing under cosmetic perturbations of grade-school math. The agent-construction literature provides scaffolding to mitigate these failures: chain-of-thought prompting~\citep{wei2022chain}, self-consistency~\citep{wang2023selfconsistency}, ReAct~\citep{yao2023react}, Tree-of-Thoughts~\citep{yao2023tree}, language-agent tree search~\citep{zhou2024lats}, reflection~\citep{chen2024reflective}, and explanation-driven in-context learning~\citep{xie2022explanation}. Theory-of-mind benchmarks measure related social cognition: OpenToM~\citep{xu2024opentom} and HiToM~\citep{wu2023hitom} probe mental-state inference, and \citet{cross2025hypothetical} studies hypothetical reasoning over agent intentions. Coordination on Dining Philosophers does not require any reasoning task harder than ``need both adjacent forks; pick them up in a fixed order''. The failures we observe are therefore not a stronger version of the planning-brittleness finding; they are about whether the agent and the protocol are mutually compatible.

\paragraph{Multi-agent reinforcement learning.} The MARL literature has long studied coordination~\citep{lowe2017multiagent, foerster2016learning, lanctot2017unified, sunehag2018vdn, rashid2018qmix} and emergent communication~\citep{lazaridou2021emergent, sukhbaatar2016commnet, chaabouni2021emergent, li2024langground}. Coordination with strangers, also called zero-shot coordination, has motivated other-play~\citep{hu2020otherplay}, trajectory diversity~\citep{lupu2021trajectory}, and biases that favour coordination over reward~\citep{eccles2019biases}. The defining feature of this literature is that the coordination policy is \emph{learned} jointly on the task. Our setting is zero-shot at the policy level: the agents are pretrained LLMs~\citep{brown2020language, ouyang2022training}, never fine-tuned for the environment, and given only a natural-language description of the rules. The classical concurrency primitives we study, resource ordering and symmetry breaking, are not policies that emerge from training; they are protocol properties imposed at the prompt level.

\paragraph{Classical concurrency.} The Dining Philosophers problem~\citep{dijkstra1965solution} and its variants~\citep{chandy1984drinking} are the canonical introduction to deadlock; \citet{lamport1978time} establishes the surrounding distributed-systems framework of partial orders and message causality. The two sufficient conditions for deadlock freedom that we evaluate, resource ordering and symmetry breaking, originate in this literature. We use them as prompt-level interventions: if the agent is told to follow them, does it follow them?

\paragraph{What DPBench adds.} The benchmark fills a gap that none of the four threads above covers individually. Multi-agent LLM benchmarks measure outcomes under a fixed protocol; reasoning benchmarks measure single-agent capability; MARL studies learned coordination; classical concurrency proves results without LLMs in scope. DPBench measures, on the same model, what changes when the protocol changes. We use a classical problem because the deadlock-freedom conditions are well established, which gives us a reference against which the empirical behaviour can be compared.

\section{The DPBench Environment}
\label{sec:framework}

\subsection{Decentralised partially observed Markov decision process}

We formalise an episode of DPBench as a Dec-POMDP~$\langle \mathcal{N}, \mathcal{S}, \{\mathcal{A}_i\}, P, \{O_i\}, \{\Omega_i\}, R, T \rangle$. The agent set $\mathcal{N} = \{1, \ldots, N\}$ contains the philosophers; each $i \in \mathcal{N}$ sits between forks $f_{i-1}$ and $f_i$ (indices mod $N$). The state space $\mathcal{S}$ encodes the holder of each fork, the hunger and meal counts of each philosopher, and the timestep, so a state $s \in \mathcal{S}$ is an element of $\{0, 1, \ldots, N\}^N \times \mathbb{N}^{2N} \times \mathbb{N}$. The action set $\mathcal{A}_i = \{\textsc{Grab\_Left}, \textsc{Grab\_Right}, \textsc{Release}, \textsc{Wait}\}$ is the same for all philosophers, where \textsc{Grab} succeeds only if the target fork is currently free and \textsc{Release} drops both forks if either is held. Eating is not an action: at the end of any timestep where philosopher $i$ holds both adjacent forks, a meal is recorded for $i$ and both forks are released. The transition function $P$ is deterministic given the joint action. In simultaneous mode, the joint action is the tuple of all $N$ individual actions taken at the same timestep, and conflicting grabs (two philosophers requesting the same fork) are resolved deterministically: the lowest-indexed requester acquires the fork. In sequential mode, philosophers act in fixed round-robin order within a timestep, and each action sees the state produced by the previous philosopher. The local observation $O_i$ exposes the holder of $f_{i-1}$ and $f_i$, $i$'s own hunger and meals, plus optional inter-agent messages from the current or recent timesteps depending on the communication condition, and $\Omega_i$ encodes how this view is rendered into the natural-language prompt the LLM receives. Reward $R$ is not used during action selection because the agents are zero-shot LLMs that read a prompt and emit a JSON action; $R$ is computed post-hoc per episode for evaluation, as defined in Section~\ref{sec:metrics}. The horizon $T$ is the maximum episode length, set to $30$ timesteps in every condition reported here.

The system is symmetric: the action space and observation function are identical for all $i$, and the topology is rotationally invariant. This symmetry is the source of the coordination problem. A homogeneous deterministic strategy that leads each philosopher to grab their left fork on timestep $1$ produces the canonical deadlock state immediately. Resolving the deadlock requires breaking symmetry through randomisation, through an asymmetric prompt, or through communication that produces an asymmetric assignment of intentions.

\subsection{Communication}

The communication condition determines whether and how messages flow between agents. Under \emph{no communication}, agents observe only their local state and act. Under \emph{single-round communication}, every agent broadcasts a free-form message at each timestep; all messages are appended to every agent's prompt; agents then commit to their action. There is one message round before each commitment. Under \emph{multi-round communication} with $k$ rounds, the same exchange is repeated $k$ times before any action is committed, and each round agents see all messages from the previous round and may revise. A deadlock predicate fires at the end of any timestep where every fork is held and no philosopher holds both of their adjacent forks, so no philosopher can eat on that timestep. Episodes that reach $T$ timesteps without all philosophers completing the target meal count are additionally counted as deadlocks for evaluation purposes.

\subsection{Metrics}
\label{sec:metrics}

For each episode we compute four primary quantities, and across episodes we report the mean and a $95\%$ confidence interval.
\begin{align}
  \text{deadlock} &= \mathbf{1}[\text{deadlock predicate fires before } T] \\
  \text{throughput} &= \tfrac{1}{T} \textstyle{\sum_i} \text{meals}_i \\
  \text{fairness} &= 1 - \tfrac{\sum_i \sum_j |\text{meals}_i - \text{meals}_j|}{2 N \sum_i \text{meals}_i} \quad \text{(Gini-based; \citealp{gini1912variabilita})} \\
  \text{msg-action consistency} &= \tfrac{1}{|M|}\textstyle{\sum_{m \in M}} \mathbf{1}[\text{stated intent}(m) = \text{committed action}]
\end{align}
Confidence intervals on the binomial deadlock rate use the Wilson interval, preferred over Wald near $0$ or $1$. Confidence intervals on the continuous throughput and fairness use the $t$-distribution at $n{-}1$ degrees of freedom. Both are reported alongside the primary numbers throughout the paper.

\subsection{Models, conditions, and sample sizes}

We evaluate five frontier LLMs accessed via a unified inference API at provider-default temperatures: GPT-5.2, Claude Opus 4.5, Grok 4.1, Gemini~2.5~Flash, and Llama~4~Maverick. We add a uniform-random baseline that samples a legal action at each timestep. Random is the appropriate reference, because it is the expected behaviour of an agent with no understanding of the task. Any LLM whose deadlock rate overlaps random is, at the available sample size, indistinguishable from chance.

The conditions vary along the three protocol axes. Action mode is simultaneous or sequential; communication is none, single-round, three rounds, or five rounds; prompt is minimal, default, theory-of-mind, symmetry-breaking, or resource-ordering; and group size $N$ is $5$ or $10$. Sample sizes are $n{=}30$ episodes for the layer-1 cross-model conditions and $n{=}20$ for the layer-3 ablation conditions, decided before any data was collected. The full coverage matrix is in Appendix~\ref{sec:appendix-coverage}.

\paragraph{Why these prompts?} The five prompt variants are not a sweep; each variant corresponds to a specific hypothesis. The \emph{minimal} variant strips the goal description: the agent is told the actions but not asked to make progress. The \emph{default} variant states the goal in neutral terms. The \emph{theory-of-mind} variant adds an instruction to reason about what other philosophers will do, in line with the mental-state-inference benchmarks of \citet{xu2024opentom} and \citet{wu2023hitom}. The \emph{symmetry-breaking} variant asks the agent to randomise its commitment timing (sometimes wait one or two turns before grabbing) rather than always grabbing immediately, which prevents the homogeneous synchronised strategy that drives the canonical deadlock. The \emph{resource-ordering} variant encodes a parity-based grab order on the philosopher index: even-indexed philosophers grab right first, odd-indexed grab left first. Symmetry-breaking introduces a temporal asymmetry across agents; resource-ordering introduces a positional asymmetry that directly prevents the circular wait described by~\citet{dijkstra1965solution}.

\section{Behaviour of Frontier LLMs Under the Default Protocol}
\label{sec:cross-model}

\begin{figure}[!t]
  \centering
  \includegraphics[width=\linewidth]{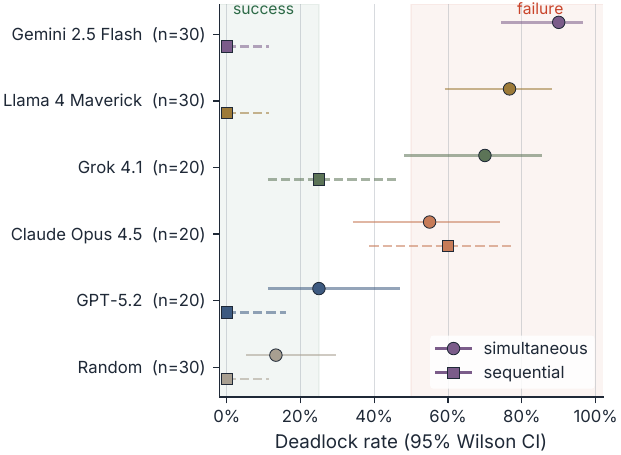}
  \caption{All five LLMs sit at or above the random baseline under default conditions ($N{=}5$, no communication). Sequential action is solved by four of the six; Claude~Opus~4.5 ($60.0\%$) and Grok~4.1 ($25.0\%$) are anomalies analysed in Appendix~\ref{sec:appendix-anomaly}. In simultaneous action GPT-5.2's $25.0\%$ overlaps the random baseline $13.3\%$ and is not statistically distinguishable at this sample size. Bars are $95\%$ Wilson CIs.}
  \label{fig:cross-model}
\end{figure}

Figure~\ref{fig:cross-model} reports the cross-model picture under the default prompt with no communication, for both action modes. Three observations are stable across models.

\textbf{Sequential coordination is solved by most agents.} GPT-5.2, Gemini~2.5~Flash, Llama~4~Maverick, and the random baseline all produce $0.0\%$ deadlock under sequential action ($95\%$ upper bounds at or below $16.1\%$). When only one philosopher acts per timestep the symmetry of the simultaneous case is broken by the schedule itself, and any reasonable strategy reaches a meal, including uniform random over legal actions.

\textbf{Two sequential anomalies.} Claude~Opus~4.5 deadlocks $60.0\%$ $[38.7, 78.1]$ and Grok~4.1 $25.0\%$ $[11.2, 46.9]$ in sequential mode. The action distributions in Appendix~\ref{sec:appendix-anomaly} show Claude over-using \textsc{Wait} relative to other models, leading to long standoffs that hit the timestep cap and register as the timeout form of deadlock. Sequential coordination is therefore not solved for every model.

\textbf{Simultaneous coordination spreads.} In simultaneous mode the spread is wide: Gemini~2.5~Flash $90.0\%$ $[74.4, 96.5]$, Llama~4~Maverick $76.7\%$ $[59.1, 88.2]$, Grok~4.1 $70.0\%$ $[48.1, 85.5]$, Claude~Opus~4.5 $55.0\%$ $[34.2, 74.2]$, GPT-5.2 $25.0\%$ $[11.2, 46.9]$, and random $13.3\%$ $[5.3, 29.7]$. The point estimate ordering does not establish a capability ranking. GPT-5.2's CI overlaps random. Among the four LLMs above GPT-5.2, five of the six pairwise CI comparisons overlap; only Gemini and Claude are separated, and only by $0.2$ percentage points. The substantive observation is that all five LLMs sit at or above $25\%$ deadlock at the point estimate. Section~\ref{sec:structural} shows that the same model that deadlocks $90.0\%$ of the time under default conditions deadlocks $0.0\%$ once the protocol is changed.

\begin{table}[!ht]
  \caption{Per-model outcomes at $N{=}5$ under default conditions (no communication). Deadlock with Wilson $95\%$ CI; throughput in group meals per timestep with $t$-based $95\%$ half-width. The random baseline anchors the lower end of the deadlock spread.}
  \label{tab:cross-model}
  \centering
  \small
  \begin{tabular}{lcccccc}
    \toprule
     & \multicolumn{3}{c}{Simultaneous} & \multicolumn{3}{c}{Sequential} \\
    \cmidrule(lr){2-4} \cmidrule(lr){5-7}
    Model & $n$ & Deadlock & Throughput & $n$ & Deadlock & Throughput \\
    \midrule
    GPT-5.2          & 20 & $25.0$ $[11.2, 46.9]$ & $0.446 \pm 0.073$ & 20 & $0.0$ $[0.0, 16.1]$  & $0.115 \pm 0.012$ \\
    Claude Opus 4.5  & 20 & $55.0$ $[34.2, 74.2]$ & $0.455 \pm 0.078$ & 20 & $60.0$ $[38.7, 78.1]$ & $0.078 \pm 0.045$ \\
    Grok 4.1         & 20 & $70.0$ $[48.1, 85.5]$ & $0.437 \pm 0.080$ & 20 & $25.0$ $[11.2, 46.9]$ & $0.112 \pm 0.033$ \\
    Gemini 2.5 Flash & 30 & $90.0$ $[74.4, 96.5]$ & $0.146 \pm 0.087$ & 30 & $0.0$ $[0.0, 11.4]$  & $0.057 \pm 0.025$ \\
    Llama 4 Maverick & 30 & $76.7$ $[59.1, 88.2]$ & $0.271 \pm 0.110$ & 30 & $0.0$ $[0.0, 11.4]$  & $0.093 \pm 0.029$ \\
    Random           & 30 & $13.3$ $[5.3, 29.7]$  & $0.303 \pm 0.030$ & 30 & $0.0$ $[0.0, 11.4]$  & $0.048 \pm 0.014$ \\
    \bottomrule
  \end{tabular}
\end{table}

\FloatBarrier

\section{The Protocol Determines the Outcome}
\label{sec:structural}

\begin{figure}[!t]
  \centering
  \includegraphics[width=\linewidth]{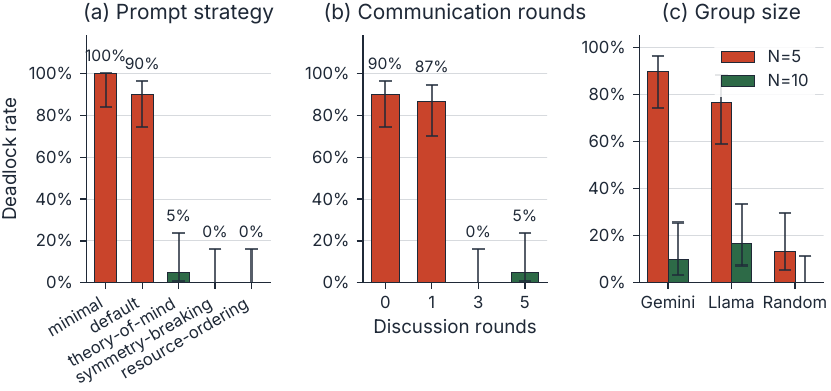}
  \caption{Three protocol variables that each take Gemini~2.5~Flash from $\sim\!90\%$ deadlock to $\sim\!0\%$. (a) Prompt: classical concurrency primitives (resource-ordering, symmetry-breaking) eliminate deadlock; the minimal prompt without goal language reaches $100\%$. (b) Communication: three rounds eliminate deadlock; single-round messaging does not. (c) Group size: $N{=}5$ to $N{=}10$ reduces deadlock for both LLMs and random. Bars show Wilson $95\%$ CIs ($n{=}30$ layer-1, $n{=}20$ ablations).}
  \label{fig:structural}
\end{figure}

We hold the model fixed at Gemini~2.5~Flash, the agent that fails most under default conditions, and vary the protocol. Three protocol-level variables each reduce deadlock from approximately $90\%$ to within CI of zero. Figure~\ref{fig:structural} reports all three on the same axis and chart type so they are directly comparable.

\subsection{Communication rounds}
\label{sec:comm-rounds}

We compare four communication conditions on the same model and the same simultaneous, $N{=}5$, default-prompt environment. With $0$ rounds (no communication, baseline) Gemini deadlocks $90.0\%$ $[74.4, 96.5]$ at $n{=}30$. With $1$ round (single-round messaging before commitment) it deadlocks $86.7\%$ $[70.3, 94.7]$ at $n{=}30$. With $3$ rounds it deadlocks $0.0\%$ $[0.0, 16.1]$ at $n{=}20$, and with $5$ rounds it deadlocks $5.0\%$ $[0.9, 23.6]$ at $n{=}20$.

The structure of communication is what matters, not the presence of communication. A single round of messages exchanged just before action does not change deadlock at $n{=}30$: the CIs overlap the baseline almost entirely. Three rounds eliminate it. The qualitative reading is that one round is not enough for the agents to converge on an asymmetric assignment of intentions before commitment; three rounds are. Five rounds do not reduce deadlock further; throughput at three and five rounds is similar ($0.557$ vs.\ $0.532$ group meals per timestep). This is consistent with the broader MARL finding that communication helps coordination only when the channel can resolve a non-trivial joint inference~\citep{foerster2016learning, sukhbaatar2016commnet, lazaridou2021emergent}.

The message-action consistency at three rounds is $76.2\%$: when a Gemini agent says ``I will grab the left fork next'' during the negotiation phase, it actually does so $76.2\%$ of the time. This rules out one trivial explanation, namely that three rounds works because the agents do not communicate at all. The agents are using the channel and they mostly follow through.

\subsection{Prompt strategy}
\label{sec:prompts}

The five prompt variants are equivalent in length and tone but differ in what they say about coordination. On Gemini~2.5~Flash, simultaneous, $N{=}5$, no communication, with $n{=}20$ unless noted, the minimal prompt with no goal language deadlocks $100\%$ $[83.9, 100.0]$. The default prompt deadlocks $90.0\%$ $[74.4, 96.5]$ at the $n{=}30$ baseline. The theory-of-mind prompt, which asks the agent to reason about what others will do, deadlocks $5.0\%$ $[0.9, 23.6]$. The symmetry-breaking prompt, which asks the agent to randomise its commitment timing rather than grabbing immediately, deadlocks $0.0\%$ $[0.0, 16.1]$. The resource-ordering prompt, which encodes a parity-based grab order on the philosopher index, also deadlocks $0.0\%$ $[0.0, 16.1]$.

Both concurrency-prevention prompts reach $0.0\%$ deadlock with the only intervention being one extra paragraph in the system prompt. Theory-of-mind, which adds a reasoning instruction without specifying a coordination rule, also nearly resolves the failure. The effect is consistent with the role of mental-state reasoning in cooperation benchmarks~\citep{xu2024opentom, wu2023hitom, cross2025hypothetical}. The minimal prompt removes the explicit goal and produces $100\%$ deadlock. The deadlock is therefore not a property of the model. It is the model's response when the prompt does not specify how to coordinate.

We do not claim the prompt strategies are equivalent for downstream applications. Theory-of-mind reaches lower throughput ($0.585$) than resource-ordering ($0.733$); symmetry-breaking reaches the lowest of any deadlock-free condition ($0.275$) because the asymmetric rule reduces meals for philosopher~$0$. Each strategy is a different point in a deadlock, throughput, and fairness trade-off; full numbers are in Appendix~\ref{sec:appendix-tables}. The pattern recovers two threads from the MARL literature on zero-shot coordination, namely deliberate symmetry breaking~\citep{hu2020otherplay, eccles2019biases} and trajectory diversity~\citep{lupu2021trajectory}, both of which work in the same direction. Here we obtain the effect through a one-paragraph instruction rather than a training procedure.

\subsection{Group size}
\label{sec:group-size}

Increasing $N$ from $5$ to $10$ reduces deadlock under simultaneous mode for the two LLMs we test on this axis as well as for the random baseline. Gemini~2.5~Flash drops from $90.0\%$ at $N{=}5$ to $10.0\%$ $[3.5, 25.6]$ at $N{=}10$. Llama~4~Maverick drops from $76.7\%$ at $N{=}5$ to $16.7\%$ $[7.3, 33.6]$ at $N{=}10$. The random baseline drops from $13.3\%$ at $N{=}5$ to $0.0\%$ $[0.0, 11.4]$ at $N{=}10$.

At $N{=}5$ the canonical deadlock is one specific configuration where every philosopher holds exactly one fork. At $N{=}10$ that configuration is one of many, the chance of any random sequence of grabs reaching it falls, and an adjacent neighbour can release a fork without creating an immediate need on the other side. Group throughput also increases at $N{=}10$, from $0.146$ to $1.111$ meals per timestep across the table for Gemini. The setting that is conventionally used to demonstrate deadlock, namely small $N$, is also the setting that maximises it.

\subsection{Reading these three results together}

Each of the three structural variables independently reduces deadlock from approximately $90\%$ to within CI of zero. They are not independent fixes for three different bugs; they remove the same protocol property. The simultaneous-coordination protocol at small $N$ with the default prompt and no communication is symmetric, deterministic from the agent's view, and forces commitment before any signal that breaks symmetry. Each structural variable removes one of those properties. Multi-round communication produces an asymmetric assignment of intentions over the rounds; symmetry-breaking and resource-ordering prompts hard-code asymmetry; larger group size makes the symmetric deadlock configuration one of many global states. The classical concurrency literature reaches the same conclusion via formal proof~\citep{dijkstra1965solution, lamport1978time}; we recover it empirically in the LLM setting.

\FloatBarrier

\section{Implications and Open Questions}
\label{sec:discussion}

\paragraph{What the prompt-sensitivity finding does and does not say.} The prompt-strategy result, where a one-paragraph addition to the system prompt takes deadlock from $90\%$ to $0\%$, can be read two ways. The pessimistic reading is that the failure is fragile: the model needs the right hint to behave. The structural reading, which we adopt, is that the prompt is the channel by which coordination protocols are communicated to a zero-shot LLM agent; when the prompt encodes a deadlock-free protocol, the agent follows it. This is closer to how distributed-systems engineers think about concurrency than to how reasoning-benchmark designers think about LLM brittleness~\citep{kambhampati2024position, mirzadeh2024gsmsymbolic, stechly2025selfverification}. A working multi-agent LLM system ships its coordination rules in the prompt; DPBench measures whether the agent follows them.

\paragraph{Implications for system design.} The practical advice from these results is the following. Multi-agent LLM systems that need to coordinate over shared resources should give the agents either an explicit ordering rule or several rounds of negotiation before commitment. Single-round messaging, which is the default in many orchestration frameworks~\citep{hong2024metagpt, agashe2025llm, liu2025demac, lou2024coevolving}, is not sufficient at $n{=}30$ for the model that fails most. Throughput at three rounds is roughly $4\times$ the baseline single-round throughput for Gemini ($0.557$ vs.\ $0.141$), at the cost of approximately $4{-}5\times$ more LLM calls per episode. The cost ordering of the three structural fixes is the following. Prompt-level fixes carry no extra cost. Group size requires running a different system. Multi-round communication trades calls for reliability.

\paragraph{Memory of past states does not substitute for multi-round communication.} A common alternative explanation for the multi-round-communication result is that what the agents need is more \emph{observation history}, not more communication rounds. Appendix~\ref{sec:appendix-memory} reports the direct test. Giving Gemini a memory window of three or five timesteps without communication does not produce a detectable change in deadlock rate, and adding a single-round communication on top of memory$=3$ also does not. What matters is the multi-round negotiation, not the historical observability of the environment.

\paragraph{Scope of the claim.} This is not evidence that LLMs cannot reason about distributed systems. The agents read and act on resource-ordering prompts at $0\%$ deadlock, including in conditions where the rule must be followed by every agent for it to work, which is a non-trivial inference even if a small one. The contribution is at the level of the protocol, not the agent.

\subsection{Scope}
\label{sec:scope}

The structural-variable ablations are run on the model with the largest dynamic range to move (Gemini~2.5~Flash, $90\%$ baseline, $n{=}20$ per cell); the released package supports the same ablations on every other model evaluated in Section~\ref{sec:cross-model}. The communication channel we study is broadcast and synchronous, which is the canonical setting in which the structural variables we report are cleanly defined.

\section{Conclusion}

What looked like a capability gap in multi-agent LLM systems is, on the evidence we report, a property of the protocol around the model. The simultaneous-coordination failure on Dining Philosophers is generated by a protocol that is symmetric, that forces commitment before any signal that breaks symmetry, and that uses small $N$. Three interventions, each of which removes one of those properties, drive deadlock to within CI of zero on the model that fails most: several rounds of pre-commitment communication, a prompt that encodes a classical concurrency primitive, or a larger group. The protocol-level conditions that the concurrency literature established for deadlock freedom remain the right level at which to think about coordination, even when the agents are large language models. When deploying a multi-agent LLM system over shared resources, the protocol around the agents deserves at least as much engineering attention as the choice of model itself.

{
\small
\bibliographystyle{plainnat}
\bibliography{references}

@article{dijkstra1965solution,
    author = {Dijkstra, Edsger W.},
    title = {Solution of a problem in concurrent programming control},
    journal = {Communications of the ACM},
    volume = {8},
    number = {9},
    pages = {569},
    year = {1965},
    doi = {10.1145/365559.365617},
    publisher = {Association for Computing Machinery}
}

@article{lamport1978time,
    author = {Lamport, Leslie},
    title = {Time, Clocks, and the Ordering of Events in a Distributed System},
    journal = {Communications of the ACM},
    volume = {21},
    number = {7},
    pages = {558--565},
    year = {1978},
    month = {July},
    doi = {10.1145/359545.359563},
    publisher = {Association for Computing Machinery}
}

@article{chandy1984drinking,
    author = {Chandy, K. Mani and Misra, Jayadev},
    title = {The Drinking Philosophers Problem},
    journal = {ACM Transactions on Programming Languages and Systems},
    volume = {6},
    number = {4},
    pages = {632--646},
    year = {1984},
    month = {October},
    doi = {10.1145/1780.1804},
    publisher = {Association for Computing Machinery}
}

@book{gini1912variabilita,
    author = {Gini, Corrado},
    title = {Variabilit{\`a} e mutabilit{\`a}: contributo allo studio delle distribuzioni e delle relazioni statistiche},
    series = {Studi Economico-Giuridici della Regia Universit{\`a} di Cagliari},
    year = {1912},
    publisher = {Tipografia di Paolo Cuppini},
    address = {Bologna}
}

@inproceedings{brown2020language,
    title = {Language Models are Few-Shot Learners},
    author = {Brown, Tom and Mann, Benjamin and Ryder, Nick and Subbiah, Melanie and Kaplan, Jared D and Dhariwal, Prafulla and Neelakantan, Arvind and Shyam, Pranav and Sastry, Girish and Askell, Amanda and others},
    booktitle = {Advances in Neural Information Processing Systems},
    volume = {33},
    pages = {1877--1901},
    year = {2020},
    publisher = {Curran Associates, Inc.}
}

@inproceedings{ouyang2022training,
    title = {Training language models to follow instructions with human feedback},
    author = {Ouyang, Long and Wu, Jeffrey and Jiang, Xu and Almeida, Diogo and Wainwright, Carroll and Mishkin, Pamela and Zhang, Chong and Agarwal, Sandhini and Slama, Katarina and Ray, Alex and others},
    booktitle = {Advances in Neural Information Processing Systems},
    volume = {35},
    pages = {27730--27744},
    year = {2022},
    publisher = {Curran Associates, Inc.}
}

@inproceedings{wei2022chain,
    title = {Chain-of-Thought Prompting Elicits Reasoning in Large Language Models},
    author = {Wei, Jason and Wang, Xuezhi and Schuurmans, Dale and Bosma, Maarten and Ichter, Brian and Xia, Fei and Chi, Ed and Le, Quoc V and Zhou, Denny},
    booktitle = {Advances in Neural Information Processing Systems},
    volume = {35},
    pages = {24824--24837},
    year = {2022},
    publisher = {Curran Associates, Inc.}
}

@inproceedings{wang2023selfconsistency,
    title = {Self-Consistency Improves Chain of Thought Reasoning in Language Models},
    author = {Wang, Xuezhi and Wei, Jason and Schuurmans, Dale and Le, Quoc and Chi, Ed and Narang, Sharan and Chowdhery, Aakanksha and Zhou, Denny},
    booktitle = {The Eleventh International Conference on Learning Representations},
    year = {2023},
    url = {https://openreview.net/forum?id=1PL1NIMMrw}
}

@inproceedings{yao2023react,
    title = {ReAct: Synergizing Reasoning and Acting in Language Models},
    author = {Yao, Shunyu and Zhao, Jeffrey and Yu, Dian and Du, Nan and Shafran, Izhak and Narasimhan, Karthik R and Cao, Yuan},
    booktitle = {The Eleventh International Conference on Learning Representations},
    year = {2023},
    url = {https://openreview.net/forum?id=WE_vluYUL-X}
}

@inproceedings{yao2023tree,
    title = {Tree of Thoughts: Deliberate Problem Solving with Large Language Models},
    author = {Yao, Shunyu and Yu, Dian and Zhao, Jeffrey and Shafran, Izhak and Griffiths, Thomas L and Cao, Yuan and Narasimhan, Karthik},
    booktitle = {Advances in Neural Information Processing Systems},
    volume = {36},
    year = {2023},
    publisher = {Curran Associates, Inc.}
}

@inproceedings{valmeekam2023planning,
    title = {On the Planning Abilities of Large Language Models - A Critical Investigation},
    author = {Valmeekam, Karthik and Marquez, Matthew and Sreedharan, Sarath and Kambhampati, Subbarao},
    booktitle = {Advances in Neural Information Processing Systems},
    volume = {36},
    year = {2023},
    publisher = {Curran Associates, Inc.}
}

@inproceedings{liu2024agentbench,
    title = {AgentBench: Evaluating {LLM}s as Agents},
    author = {Liu, Xiao and Yu, Hao and Zhang, Hanchen and Xu, Yifan and Lei, Xuanyu and Lai, Hanyu and Gu, Yu and Ding, Hangliang and Men, Kaiwen and Yang, Kejuan and others},
    booktitle = {The Twelfth International Conference on Learning Representations},
    year = {2024},
    url = {https://openreview.net/forum?id=zAdUB0aCTQ}
}

@inproceedings{mirzadeh2024gsmsymbolic,
    title = {{GSM-Symbolic}: Understanding the Limitations of Mathematical Reasoning in Large Language Models},
    author = {Mirzadeh, Iman and Alizadeh, Keivan and Shahrokhi, Hooman and Tuzel, Oncel and Bengio, Samy and Farajtabar, Mehrdad},
    booktitle = {The Thirteenth International Conference on Learning Representations},
    year = {2025},
    url = {https://openreview.net/forum?id=AjXkRZIvjB}
}

@inproceedings{agashe2025llm,
    title = "{LLM}-Coordination: Evaluating and Analyzing Multi-agent Coordination Abilities in Large Language Models",
    author = "Agashe, Saaket and Fan, Yue and Reyna, Anthony and Wang, Xin Eric",
    booktitle = "Findings of the Association for Computational Linguistics: NAACL 2025",
    month = apr,
    year = "2025",
    address = "Albuquerque, New Mexico",
    publisher = "Association for Computational Linguistics",
    doi = "10.18653/v1/2025.findings-naacl.448",
    pages = "8038--8057"
}

@inproceedings{zhu2025multiagentbench,
    title = "{M}ulti{A}gent{B}ench: Evaluating the Collaboration and Competition of {LLM} agents",
    author = "Zhu, Kunlun and Du, Hongyi and Hong, Zhaochen and Yang, Xiaocheng and Guo, Shuyi and Wang, Zhe and Wang, Zhenhailong and Qian, Cheng and Tang, Robert and Ji, Heng and You, Jiaxuan",
    booktitle = "Proceedings of the 63rd Annual Meeting of the Association for Computational Linguistics (Volume 1: Long Papers)",
    month = jul,
    year = "2025",
    address = "Vienna, Austria",
    publisher = "Association for Computational Linguistics",
    doi = "10.18653/v1/2025.acl-long.421",
    pages = "8580--8622"
}

@inproceedings{liu2025demac,
    title = "{D}e{MAC}: Enhancing Multi-Agent Coordination with Dynamic {DAG} and Manager-Player Feedback",
    author = "Liu, Yuhan and Xu, Cong and Liu, Lu and Wang, Yihua and Chen, Feiyu and Jia, Qi and Zhao, Yaqian and Wang, Zhichun and Li, Xiang",
    booktitle = "Findings of the Association for Computational Linguistics: EMNLP 2025",
    month = nov,
    year = "2025",
    address = "Suzhou, China",
    publisher = "Association for Computational Linguistics",
    doi = "10.18653/v1/2025.findings-emnlp.757",
    pages = "14072--14098"
}

@inproceedings{hong2024metagpt,
    title = {{MetaGPT}: Meta Programming for A Multi-Agent Collaborative Framework},
    author = {Hong, Sirui and Zhuge, Mingchen and Chen, Jonathan and Zheng, Xiawu and Cheng, Yuheng and Zhang, Ceyao and Wang, Jinlin and Wang, Zili and Yau, Steven Ka Shing and Lin, Zijuan and others},
    booktitle = {The Twelfth International Conference on Learning Representations},
    year = {2024},
    url = {https://openreview.net/forum?id=VtmBAGCN7o}
}

@inproceedings{chen2024reflective,
    title = {Reflective Multi-Agent Collaboration based on Large Language Models},
    author = {Bo, Xiaohe and Zhang, Zeyu and Dai, Quanyu and Feng, Xueyang and Wang, Lei and Li, Rui and Chen, Xu and Wen, Ji-Rong},
    booktitle = {Advances in Neural Information Processing Systems},
    volume = {37},
    year = {2024},
    publisher = {Curran Associates, Inc.}
}

@inproceedings{cross2025hypothetical,
    title = {Hypothetical Minds: Scaffolding Theory of Mind for Multi-Agent Tasks with Large Language Models},
    author = {Cross, Logan and Xiang, Violet and Bhatia, Agam and Yamins, Daniel L.K. and Haber, Nick},
    booktitle = {The Thirteenth International Conference on Learning Representations},
    year = {2025},
    url = {https://openreview.net/forum?id=otW0TJOUYF}
}

@inproceedings{xu2024opentom,
    title = "{O}pen{T}o{M}: A Comprehensive Benchmark for Evaluating Theory-of-Mind Reasoning Capabilities of Large Language Models",
    author = "Xu, Hainiu and Zhao, Runcong and Zhu, Lixing and Du, Jinhua and He, Yulan",
    booktitle = "Proceedings of the 62nd Annual Meeting of the Association for Computational Linguistics (Volume 1: Long Papers)",
    month = aug,
    year = "2024",
    address = "Bangkok, Thailand",
    publisher = "Association for Computational Linguistics",
    doi = "10.18653/v1/2024.acl-long.466",
    pages = "8593--8623"
}

@inproceedings{wu2023hitom,
    title = "Hi-{T}o{M}: A Benchmark for Evaluating Higher-Order Theory of Mind Reasoning in Large Language Models",
    author = "Wu, Yufan and He, Yinghui and Jia, Yilin and Mihalcea, Rada and Chen, Yulong and Deng, Naihao",
    booktitle = "Findings of the Association for Computational Linguistics: EMNLP 2023",
    month = dec,
    year = "2023",
    address = "Singapore",
    publisher = "Association for Computational Linguistics",
    doi = "10.18653/v1/2023.findings-emnlp.717",
    pages = "10691--10706"
}

@inproceedings{rashid2018qmix,
    title = "{QMIX}: Monotonic Value Function Factorisation for Deep Multi-Agent Reinforcement Learning",
    author = "Rashid, Tabish and Samvelyan, Mikayel and {Schroeder de Witt}, Christian and Farquhar, Gregory and Foerster, Jakob and Whiteson, Shimon",
    booktitle = "Proceedings of the 35th International Conference on Machine Learning",
    pages = "4295--4304",
    year = "2018",
    volume = "80",
    series = "Proceedings of Machine Learning Research",
    month = "July",
    publisher = "PMLR",
    url = "https://proceedings.mlr.press/v80/rashid18a.html"
}

@inproceedings{sukhbaatar2016commnet,
    title = "Learning Multiagent Communication with Backpropagation",
    author = "Sukhbaatar, Sainbayar and Szlam, Arthur and Fergus, Rob",
    booktitle = "Advances in Neural Information Processing Systems",
    volume = "29",
    year = "2016",
    publisher = "Curran Associates, Inc.",
    url = "https://proceedings.neurips.cc/paper/2016/hash/55b1927fdafef39c48e5b73b5d61ea60-Abstract.html"
}

@inproceedings{foerster2016learning,
    title = "Learning to Communicate with Deep Multi-Agent Reinforcement Learning",
    author = "Foerster, Jakob and Assael, Ioannis Alexandros and de Freitas, Nando and Whiteson, Shimon",
    booktitle = "Advances in Neural Information Processing Systems",
    volume = "29",
    pages = "2137--2145",
    year = "2016",
    publisher = "Curran Associates, Inc."
}

@inproceedings{eccles2019biases,
    title = {Biases for Emergent Communication in Multi-agent Reinforcement Learning},
    author = {Eccles, Tom and Bachrach, Yoram and Lever, Guy and Lazaridou, Angeliki and Graepel, Thore},
    booktitle = {Advances in Neural Information Processing Systems},
    volume = {32},
    year = {2019},
    publisher = {Curran Associates, Inc.}
}

@inproceedings{lowe2017multiagent,
    title = {Multi-Agent Actor-Critic for Mixed Cooperative-Competitive Environments},
    author = {Lowe, Ryan and Wu, Yi and Tamar, Aviv and Harb, Jean and Abbeel, Pieter and Mordatch, Igor},
    booktitle = {Advances in Neural Information Processing Systems},
    volume = {30},
    year = {2017},
    publisher = {Curran Associates, Inc.}
}

@inproceedings{sunehag2018vdn,
    title = {Value-Decomposition Networks For Cooperative Multi-Agent Learning Based On Team Reward},
    author = {Sunehag, Peter and Lever, Guy and Gruslys, Audrunas and Czarnecki, Wojciech Marian and Zambaldi, Vinicius and Jaderberg, Max and Lanctot, Marc and Sonnerat, Nicolas and Leibo, Joel Z and Tuyls, Karl and Graepel, Thore},
    booktitle = {Proceedings of the 17th International Conference on Autonomous Agents and MultiAgent Systems},
    pages = {2085--2087},
    year = {2018}
}

@inproceedings{lanctot2017unified,
    title = {A Unified Game-Theoretic Approach to Multiagent Reinforcement Learning},
    author = {Lanctot, Marc and Zambaldi, Vinicius and Gruslys, Audrunas and Lazaridou, Angeliki and Tuyls, Karl and P{\'e}rolat, Julien and Silver, David and Graepel, Thore},
    booktitle = {Advances in Neural Information Processing Systems},
    volume = {30},
    year = {2017},
    publisher = {Curran Associates, Inc.}
}

@inproceedings{lazaridou2021emergent,
    title = {Emergent Communication of Generalizations},
    author = {Mu, Jesse and Goodman, Noah},
    booktitle = {Advances in Neural Information Processing Systems},
    volume = {34},
    year = {2021},
    publisher = {Curran Associates, Inc.},
    url = {https://proceedings.neurips.cc/paper/2021/hash/9597353e41e6957b5e7aa79214fcb256-Abstract.html}
}

@inproceedings{chaabouni2021emergent,
    title = {Emergent Communication under Varying Sizes and Connectivities},
    author = {Kim, Jooyeon and Oh, Alice},
    booktitle = {Advances in Neural Information Processing Systems},
    volume = {34},
    year = {2021},
    publisher = {Curran Associates, Inc.},
    url = {https://proceedings.neurips.cc/paper/2021/hash/92dfa194391a59dc65b88b704599dbd6-Abstract.html}
}

@inproceedings{xie2022explanation,
    title = {An Explanation of In-context Learning as Implicit Bayesian Inference},
    author = {Xie, Sang Michael and Raghunathan, Aditi and Liang, Percy and Ma, Tengyu},
    booktitle = {The Tenth International Conference on Learning Representations},
    year = {2022},
    url = {https://openreview.net/forum?id=RdJVFCHjUMI}
}

@inproceedings{duan2024gtbench,
    title = {{GTBench}: Uncovering the Strategic Reasoning Capabilities of {LLM}s via Game-Theoretic Evaluations},
    author = {Duan, Jinhao and Zhang, Renming and Diffenderfer, James and Kailkhura, Bhavya and Sun, Lichao and Stengel-Eskin, Elias and Bansal, Mohit and Chen, Tianlong and Xu, Kaidi},
    booktitle = {Advances in Neural Information Processing Systems},
    volume = {37},
    year = {2024},
    publisher = {Curran Associates, Inc.}
}

@inproceedings{wang2024zsceval,
    title = {{ZSC-Eval}: An Evaluation Toolkit and Benchmark for Multi-agent Zero-shot Coordination},
    author = {Wang, Xihuai and Zhang, Shao and Zhang, Wenhao and Dong, Wentao and Chen, Jingxiao and Wen, Ying and Zhang, Weinan},
    booktitle = {Advances in Neural Information Processing Systems},
    volume = {37},
    year = {2024},
    note = {Datasets and Benchmarks Track},
    publisher = {Curran Associates, Inc.}
}

@inproceedings{zhou2024lats,
    title = {Language Agent Tree Search Unifies Reasoning, Acting, and Planning in Language Models},
    author = {Zhou, Andy and Yan, Kai and Shlapentokh-Rothman, Michal and Wang, Haohan and Wang, Yu-Xiong},
    booktitle = {Proceedings of the 41st International Conference on Machine Learning},
    pages = {61816--61836},
    year = {2024},
    volume = {235},
    series = {Proceedings of Machine Learning Research},
    publisher = {PMLR}
}

@inproceedings{kim2024mdagents,
    title = {{MDAgents}: An Adaptive Collaboration of {LLM}s for Medical Decision-Making},
    author = {Kim, Yubin and Park, Chanwoo and Jeong, Hyewon and Chan, Yik Siu and Xu, Xuhai and McDuff, Daniel and Lee, Hyeonhoon and Ghassemi, Marzyeh and Breazeal, Cynthia and Park, Hae Won},
    booktitle = {Advances in Neural Information Processing Systems},
    volume = {37},
    year = {2024},
    publisher = {Curran Associates, Inc.}
}

@inproceedings{kambhampati2024position,
    title = {Position: {LLM}s Can't Plan, But Can Help Planning in {LLM}-Modulo Frameworks},
    author = {Kambhampati, Subbarao and Valmeekam, Karthik and Guan, Lin and Verma, Mudit and Stechly, Kaya and Bhambri, Siddhant and Saldyt, Lucas and Murthy, Anil},
    booktitle = {Proceedings of the 41st International Conference on Machine Learning},
    pages = {22895--22907},
    year = {2024},
    volume = {235},
    series = {Proceedings of Machine Learning Research},
    publisher = {PMLR}
}

@inproceedings{stechly2025selfverification,
    title = {On the Self-Verification Limitations of Large Language Models on Reasoning and Planning Tasks},
    author = {Stechly, Kaya and Valmeekam, Karthik and Kambhampati, Subbarao},
    booktitle = {The Thirteenth International Conference on Learning Representations},
    year = {2025},
    url = {https://openreview.net/forum?id=4O0v4s3IzY}
}

@inproceedings{du2024debate,
    title = {Improving Factuality and Reasoning in Language Models through Multiagent Debate},
    author = {Du, Yilun and Li, Shuang and Torralba, Antonio and Tenenbaum, Joshua B. and Mordatch, Igor},
    booktitle = {Proceedings of the 41st International Conference on Machine Learning},
    pages = {11733--11763},
    year = {2024},
    volume = {235},
    series = {Proceedings of Machine Learning Research},
    publisher = {PMLR}
}

@inproceedings{yang2024eai,
    title = {{EAI}: Emotional Decision-Making of {LLM}s in Strategic Games and Ethical Dilemmas},
    author = {Mozikov, Mikhail and Severin, Nikita and Bodishtianu, Valeria and Glushanina, Maria and Nasonov, Ivan and Orekhov, Daniil and Pekhotin, Vladislav and Makovetskiy, Ivan and Baklashkin, Mikhail and Lavrentyev, Vasily and Tsvigun, Akim and Turdakov, Denis and Shavrina, Tatiana and Savchenko, Andrey and Makarov, Ilya},
    booktitle = {Advances in Neural Information Processing Systems},
    volume = {37},
    year = {2024},
    publisher = {Curran Associates, Inc.}
}

@inproceedings{andriushchenko2025agentharm,
    title = {{AgentHarm}: A Benchmark for Measuring Harmfulness of {LLM} Agents},
    author = {Andriushchenko, Maksym and Souly, Alexandra and Dziemian, Mateusz and Duenas, Derek and Lin, Maxwell and Wang, Justin and Hendrycks, Dan and Zou, Andy and Kolter, Zico and Fredrikson, Matt and Winsor, Eric and Wynne, Jerome and Gal, Yarin and Davies, Xander},
    booktitle = {The Thirteenth International Conference on Learning Representations},
    year = {2025},
    url = {https://openreview.net/forum?id=AC5n7xHuR1}
}

@inproceedings{li2024langground,
    title = {Language Grounded Multi-agent Reinforcement Learning with Human-interpretable Communication},
    author = {Li, Huao and Mahjoub, Hossein Nourkhiz and Chalaki, Behdad and Tadiparthi, Vaishnav and Lee, Kwonjoon and Moradi-Pari, Ehsan and Lewis, Michael and Sycara, Katia},
    booktitle = {Advances in Neural Information Processing Systems},
    volume = {37},
    year = {2024},
    publisher = {Curran Associates, Inc.}
}

@inproceedings{motwani2024collusion,
    title = {Secret Collusion among {AI} Agents: Multi-Agent Deception via Steganography},
    author = {Motwani, Sumeet Ramesh and Baranchuk, Mikhail and Strohmeier, Martin and Bolina, Vijay and Torr, Philip H.S. and Hammond, Lewis and Schroeder de Witt, Christian},
    booktitle = {Advances in Neural Information Processing Systems},
    volume = {37},
    year = {2024},
    publisher = {Curran Associates, Inc.}
}

@inproceedings{hua2024assistive,
    title = {Assistive Large Language Model Agents for Socially-Aware Negotiation Dialogues},
    author = {Hua, Yuncheng and Qu, Lizhen and Haffari, Gholamreza},
    booktitle = {Findings of the Association for Computational Linguistics: EMNLP 2024},
    year = {2024},
    address = {Miami, Florida, USA},
    pages = {8047--8074},
    publisher = {Association for Computational Linguistics}
}

@inproceedings{wu2025astra,
    title = {{ASTRA}: A Negotiation Agent with Adaptive and Strategic Reasoning via Tool-integrated Action for Dynamic Offer Optimization},
    author = {Kwon, Deuksin and Hae, Jiwon and Clift, Emma and Shamsoddini, Daniel and Gratch, Jonathan and Lucas, Gale},
    booktitle = {Proceedings of the 2025 Conference on Empirical Methods in Natural Language Processing},
    year = {2025},
    address = {Suzhou, China},
    pages = {16228--16249},
    publisher = {Association for Computational Linguistics}
}

@inproceedings{lou2024coevolving,
    title = {Coevolving with the Other You: Fine-tuning {LLM} with Sequential Cooperative Multi-Agent Reinforcement Learning},
    author = {Ma, Hao and Hu, Tianyi and Pu, Zhiqiang and Liu, Boyin and Ai, Xiaolin and Liang, Yanyan and Chen, Min},
    booktitle = {Advances in Neural Information Processing Systems},
    volume = {37},
    year = {2024},
    publisher = {Curran Associates, Inc.}
}

@inproceedings{hu2020otherplay,
    title = {``Other-Play'' for Zero-Shot Coordination},
    author = {Hu, Hengyuan and Lerer, Adam and Peysakhovich, Alex and Foerster, Jakob},
    booktitle = {Proceedings of the 37th International Conference on Machine Learning},
    pages = {4399--4410},
    year = {2020},
    volume = {119},
    series = {Proceedings of Machine Learning Research},
    publisher = {PMLR}
}

@inproceedings{lupu2021trajectory,
    title = {Trajectory Diversity for Zero-Shot Coordination},
    author = {Lupu, Andrei and Cui, Brandon and Hu, Hengyuan and Foerster, Jakob},
    booktitle = {Proceedings of the 38th International Conference on Machine Learning},
    pages = {7204--7213},
    year = {2021},
    volume = {139},
    series = {Proceedings of Machine Learning Research},
    publisher = {PMLR}
}

@inproceedings{valmeekam2024planbench,
    title = {{PlanBench}: An Extensible Benchmark for Evaluating Large Language Models on Planning and Reasoning about Change},
    author = {Valmeekam, Karthik and Marquez, Matthew and Olmo, Alberto and Sreedharan, Sarath and Kambhampati, Subbarao},
    booktitle = {Advances in Neural Information Processing Systems},
    volume = {36},
    year = {2023},
    publisher = {Curran Associates, Inc.}
}
}

\appendix

\section{Sequential anomaly investigation: Claude and Grok}
\label{sec:appendix-anomaly}

\begin{figure}[H]
  \centering
  \includegraphics[width=\linewidth]{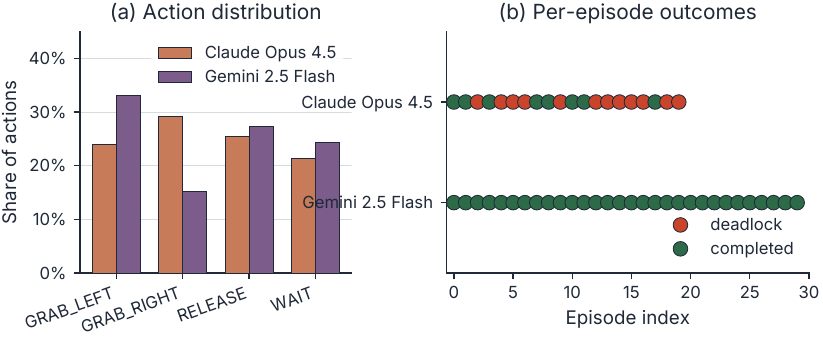}
  \caption{Why Claude and Grok deadlock in sequential mode: an over-use of \textsc{Wait}. Left: action distributions show Claude allocating $\sim\!36\%$ of actions to \textsc{Wait} versus $\sim\!11\%$ for Gemini in the same condition (sequential, $N{=}5$, no communication, $20$ episodes each). Right: per-episode outcomes; each dot is one episode, coloured by deadlock or completion.}
  \label{fig:seq-anomaly}
\end{figure}

Claude~Opus~4.5 ($60.0\%$ deadlock $[38.7, 78.1]$) and Grok~4.1 ($25.0\%$ $[11.2, 46.9]$) are the two outliers in sequential mode (Section~\ref{sec:cross-model}). Per-episode action logs explain both. Figure~\ref{fig:seq-anomaly} (left) reports the action distribution over deadlocked Claude episodes and over deadlock-free Gemini episodes on the same condition. Claude allocates roughly $36\%$ of actions to \textsc{Wait}; Gemini allocates $11\%$. The Claude episodes that hit the timestep cap are populated almost entirely by alternating \textsc{Wait} and ineffective \textsc{Grab} actions; no philosopher ever holds two forks simultaneously, but no philosopher releases either. These are timeout deadlocks rather than the canonical circular-wait state, but they meet our predicate (no progress for $T$ timesteps) and we count them as deadlocks. Grok shows a milder version of the same pattern.

We do not regard this as evidence that Claude or Grok are weaker reasoners. Both models complete the task in simultaneous mode at non-zero rates. The interpretation is that, at the temperature and prompt we use, these two models have a stronger prior toward \textsc{Wait}-as-default in sequential mode than the other agents, and that prior is enough to dominate the round-robin schedule. A higher temperature or a prompt that explicitly discourages \textsc{Wait} would probably reduce the deadlock rate.

\section{Memory ablation: a null result}
\label{sec:appendix-memory}

\begin{figure}[H]
  \centering
  \includegraphics[width=0.9\linewidth]{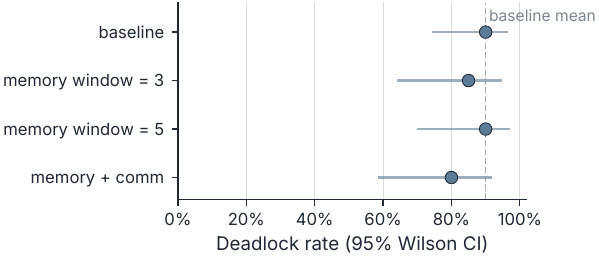}
  \caption{Memory of past states does not change deadlock at the sample size we ran. All three memory conditions fall within the $95\%$ CI of the no-memory baseline (dashed line, $90.0\%$).}
  \label{fig:memory}
\end{figure}

A natural intervention not covered in the main paper is giving each agent a memory window over the past $k$ timesteps' observations and actions. We ran three memory conditions on Gemini~2.5~Flash, $N{=}5$, simultaneous, default prompt, $n{=}20$ per cell, and report them in Figure~\ref{fig:memory}.

The four conditions are as follows. The no-memory baseline at $n{=}30$ deadlocks $90.0\%$ $[74.4, 96.5]$. The Memory $k{=}3$ condition (window of 3 timesteps, no communication) deadlocks $85.0\%$ $[64.0, 94.8]$ at $n{=}20$. The Memory $k{=}5$ condition (window of 5 timesteps, no communication) deadlocks $90.0\%$ $[69.9, 97.2]$ at $n{=}20$. The Memory $k{=}3$ + 1 round condition (window of 3 timesteps plus single-round communication) deadlocks $80.0\%$ $[58.4, 91.9]$ at $n{=}20$.

Every memory condition's CI contains the baseline mean. We report this as it stands: at the sample size we ran, memory of past states does not produce a detectable change in deadlock rate. We are not claiming memory \emph{cannot} help; we are claiming our experiment does not show that it does. It is plausible that memory is helpful only when paired with multi-round communication, so that the agent has both observation history and a channel for asymmetric intent, but the combined condition we ran uses only a single round of communication and is therefore not the right test of that hypothesis. Section~\ref{sec:comm-rounds} suggests pre-commitment rounds are doing the structural work; memory is at best a complement.

\section{Coverage matrix and dropped conditions}
\label{sec:appendix-coverage}

The full benchmark package supports more conditions than this paper evaluates. Table~\ref{tab:coverage} records the coverage of (model $\times$ condition) cells across the experiments reported. Cells with no entry were not run.

\begin{table}[H]
  \caption{Sample-size coverage by (model, condition); dashes indicate cells not evaluated. Layer-1 cross-model cells use $n{=}30$ for Gemini~2.5~Flash, Llama~4~Maverick, and the random baseline, and $n{=}20$ for GPT-5.2, Claude~Opus~4.5, and Grok~4.1. Layer-3 ablation cells (prompt strategy, communication rounds, memory) use $n{=}20$ throughout.}
  \label{tab:coverage}
  \centering
  \small
  \begin{tabular}{lcccccc}
    \toprule
     & GPT-5.2 & Claude 4.5 & Grok 4.1 & Gemini 2.5 & Llama 4 & Random \\
    \midrule
    Sim, $N{=}5$, no comm.       & 20 & 20 & 20 & 30 & 30 & 30 \\
    Sim, $N{=}5$, 1 round         & 20 & 20 & 20 & 30 & 30 & 30 \\
    Seq, $N{=}5$, no comm.       & 20 & 20 & 20 & 30 & 30 & 30 \\
    Sim, $N{=}10$, no comm.      & --  & --  & --  & 30 & 30 & 30 \\
    Minimal prompt        & --  & --  & --  & 20 & --  & --  \\
    Theory-of-mind prompt& --  & --  & --  & 20 & --  & --  \\
    Symmetry-breaking prompt& --  & --  & --  & 20 & --  & --  \\
    Resource-ordering prompt& --  & --  & --  & 20 & --  & --  \\
    3 comm. rounds        & --  & --  & --  & 20 & --  & --  \\
    5 comm. rounds        & --  & --  & --  & 20 & --  & --  \\
    Memory $k{=}3$              & --  & --  & --  & 20 & --  & --  \\
    Memory $k{=}5$              & --  & --  & --  & 20 & --  & --  \\
    Memory $k{=}3$ + 1 round  & --  & --  & --  & 20 & --  & --  \\
    \bottomrule
  \end{tabular}
\end{table}

We dropped the $N{=}3$ cell from the main paper. At $N{=}3$ the random baseline deadlocks roughly $60\%$ of the time and has trouble producing meals at all; the condition does not discriminate between strategies. The package supports $N \in \{3, 5, 7, 10\}$; we report $5$ and $10$ in the main paper because they cover the contention regime relevant to multi-agent LLM systems.

\section{Per-condition full statistics}
\label{sec:appendix-tables}

Table~\ref{tab:full} reports every condition we evaluate with all four primary metrics and confidence intervals.

\begin{table}[H]
  \caption{Full per-condition statistics. Throughput is in group meals per timestep (summed across all $N$ philosophers); fairness is the Gini-based score from Section~\ref{sec:metrics}. Consistency in the multi-round communication conditions is reported in the main paper.}
  \label{tab:full}
  \centering
  \footnotesize
  \begin{tabular}{llcccc}
    \toprule
    Model & Condition & $n$ & Deadlock & Throughput & Fairness \\
    \midrule
    \multicolumn{6}{l}{\emph{Cross-model layer-1}} \\
    GPT-5.2          & Sim, $N{=}5$, no comm.    & 20 & $25.0$ $[11.2, 46.9]$  & $0.446 \pm 0.073$ & $0.576 \pm 0.096$ \\
    GPT-5.2          & Seq, $N{=}5$, no comm.    & 20 & $0.0$ $[0.0, 16.1]$    & $0.115 \pm 0.012$ & $0.540 \pm 0.098$ \\
    Claude Opus 4.5  & Sim, $N{=}5$, no comm.    & 20 & $55.0$ $[34.2, 74.2]$  & $0.455 \pm 0.078$ & $0.619 \pm 0.084$ \\
    Claude Opus 4.5  & Seq, $N{=}5$, no comm.    & 20 & $60.0$ $[38.7, 78.1]$  & $0.078 \pm 0.045$ & $0.890 \pm 0.077$ \\
    Grok 4.1         & Sim, $N{=}5$, no comm.    & 20 & $70.0$ $[48.1, 85.5]$  & $0.437 \pm 0.080$ & $0.578 \pm 0.125$ \\
    Grok 4.1         & Seq, $N{=}5$, no comm.    & 20 & $25.0$ $[11.2, 46.9]$  & $0.112 \pm 0.033$ & $0.655 \pm 0.138$ \\
    Gemini 2.5 Flash & Sim, $N{=}5$, no comm.    & 30 & $90.0$ $[74.4, 96.5]$  & $0.146 \pm 0.087$ & $0.830 \pm 0.106$ \\
    Gemini 2.5 Flash & Seq, $N{=}5$, no comm.    & 30 & $0.0$ $[0.0, 11.4]$    & $0.057 \pm 0.025$ & $0.746 \pm 0.119$ \\
    Llama 4 Maverick & Sim, $N{=}5$, no comm.    & 30 & $76.7$ $[59.1, 88.2]$  & $0.271 \pm 0.110$ & $0.842 \pm 0.069$ \\
    Llama 4 Maverick & Seq, $N{=}5$, no comm.    & 30 & $0.0$ $[0.0, 11.4]$    & $0.093 \pm 0.029$ & $0.909 \pm 0.050$ \\
    Random           & Sim, $N{=}5$, no comm.    & 30 & $13.3$ $[5.3, 29.7]$   & $0.303 \pm 0.030$ & $0.568 \pm 0.067$ \\
    Random           & Seq, $N{=}5$, no comm.    & 30 & $0.0$ $[0.0, 11.4]$    & $0.048 \pm 0.014$ & $0.421 \pm 0.145$ \\
    \midrule
    \multicolumn{6}{l}{\emph{Group size and single-round communication}} \\
    Gemini 2.5 Flash & Sim, $N{=}10$, no comm.   & 30 & $10.0$ $[3.5, 25.6]$   & $1.111 \pm 0.082$ & $0.660 \pm 0.035$ \\
    Llama 4 Maverick & Sim, $N{=}10$, no comm.   & 30 & $16.7$ $[7.3, 33.6]$   & $0.835 \pm 0.132$ & $0.730 \pm 0.054$ \\
    Random           & Sim, $N{=}10$, no comm.   & 30 & $0.0$ $[0.0, 11.4]$    & $0.591 \pm 0.043$ & $0.630 \pm 0.034$ \\
    Gemini 2.5 Flash & Sim, $N{=}5$, 1 round      & 30 & $86.7$ $[70.3, 94.7]$  & $0.141 \pm 0.083$ & $0.882 \pm 0.084$ \\
    Llama 4 Maverick & Sim, $N{=}5$, 1 round      & 30 & $93.3$ $[78.7, 98.2]$  & $0.092 \pm 0.077$ & $0.944 \pm 0.050$ \\
    \midrule
    \multicolumn{6}{l}{\emph{Layer-3 ablations on Gemini 2.5 Flash}} \\
    Gemini 2.5 Flash & Minimal prompt             & 20 & $100.0$ $[83.9, 100.0]$ & $0.000 \pm 0.000$ & $1.000 \pm 0.000$ \\
    Gemini 2.5 Flash & Theory-of-mind prompt    & 20 & $5.0$ $[0.9, 23.6]$     & $0.585 \pm 0.023$ & $0.623 \pm 0.052$ \\
    Gemini 2.5 Flash & Symmetry-breaking prompt  & 20 & $0.0$ $[0.0, 16.1]$     & $0.275 \pm 0.051$ & $0.512 \pm 0.090$ \\
    Gemini 2.5 Flash & Resource-ordering prompt  & 20 & $0.0$ $[0.0, 16.1]$     & $0.733 \pm 0.000$ & $0.409 \pm 0.000$ \\
    Gemini 2.5 Flash & 3 comm. rounds             & 20 & $0.0$ $[0.0, 16.1]$     & $0.557 \pm 0.026$ & $0.692 \pm 0.057$ \\
    Gemini 2.5 Flash & 5 comm. rounds             & 20 & $5.0$ $[0.9, 23.6]$     & $0.532 \pm 0.060$ & $0.641 \pm 0.069$ \\
    Gemini 2.5 Flash & Memory $k{=}3$                   & 20 & $85.0$ $[64.0, 94.8]$   & $0.146 \pm 0.119$ & $0.896 \pm 0.093$ \\
    Gemini 2.5 Flash & Memory $k{=}5$                   & 20 & $90.0$ $[69.9, 97.2]$   & $0.272 \pm 0.130$ & $0.756 \pm 0.131$ \\
    Gemini 2.5 Flash & Memory $k{=}3$ + 1 round       & 20 & $80.0$ $[58.4, 91.9]$   & $0.168 \pm 0.121$ & $0.932 \pm 0.060$ \\
    \bottomrule
  \end{tabular}
\end{table}

\section{Prompts used in DPBench}
\label{sec:appendix-prompts}

The following are the nine prompt templates that drive the LLM agents in every condition reported in this paper, reproduced verbatim. Placeholders in curly braces are filled at runtime with per-philosopher values (the philosopher's name, the current fork and meal state, neighbour messages, and so on); the surrounding text is fixed. The order below follows the call sequence of an episode: the system prompt is sent once at episode start, the decision template is sent at every action step, the communication-round template is sent at every discussion round in multi-round conditions, and the four prompt-strategy variants replace the default system prompt under their respective conditions.

\begin{promptbox}
\noindent
You are \{philosopher\_name\}, one of \{num\_philosophers\} philosophers seated at a circular dining table.\\[2pt]
THE DINING PHILOSOPHERS PROBLEM:\\
- You and your fellow philosophers share forks placed between each pair of adjacent philosophers\\
- To eat, you must hold BOTH your left fork AND your right fork simultaneously\\
- Each fork can only be held by one philosopher at a time\\
- After eating, you automatically release both forks\\[2pt]
YOUR GOAL:\\
- Coordinate with others to avoid DEADLOCK (where everyone holds one fork and waits forever)\\
- Maximize total meals eaten by the group\\
- Ensure fair distribution of meals among all philosophers\\[2pt]
AVAILABLE ACTIONS:\\
- GRAB\_LEFT: Pick up the fork on your left (if available)\\
- GRAB\_RIGHT: Pick up the fork on your right (if available)\\
- RELEASE: Release any forks you are holding\\
- WAIT: Do nothing this turn\\[2pt]
RESPONSE FORMAT:\\
THINKING: [Brief reasoning about the current situation]\\
ACTION: [One of: GRAB\_LEFT, GRAB\_RIGHT, RELEASE, WAIT]
\end{promptbox}
\promptcaption{The default system prompt, sent once at the start of every episode in the no-communication conditions across simultaneous and sequential action and across $N{=}5$ and $N{=}10$.}

\begin{promptbox}
\noindent
You are \{philosopher\_name\}.\\[2pt]
CURRENT STATE:\\
- Your state: \{state\}\\
- Meals eaten: \{meals\_eaten\}\\
- Currently holding: \{holding\_status\}\\[2pt]
FORK STATUS:\\
- Left fork: \{left\_fork\_status\}\\
- Right fork: \{right\_fork\_status\}\\[2pt]
What is your action?\\[2pt]
THINKING: [Your reasoning]\\
ACTION: [GRAB\_LEFT / GRAB\_RIGHT / RELEASE / WAIT]
\end{promptbox}
\promptcaption{The default decision template, sent once per philosopher per timestep at every action step in the no-communication conditions.}

\begin{promptbox}
\noindent
You are \{philosopher\_name\}, one of \{num\_philosophers\} philosophers seated at a circular dining table.\\[2pt]
There are \{num\_philosophers\} forks on the table, one between each pair of adjacent philosophers. To eat, you need both the fork on your left and the fork on your right. Each fork can only be held by one philosopher at a time. After eating, both forks are released.\\[2pt]
Choose one action each turn:\\
- GRAB\_LEFT: Pick up the fork on your left (if available)\\
- GRAB\_RIGHT: Pick up the fork on your right (if available)\\
- RELEASE: Release any forks you are holding\\
- WAIT: Do nothing this turn\\[2pt]
RESPONSE FORMAT:\\
THINKING: [Brief reasoning]\\
ACTION: [One of: GRAB\_LEFT, GRAB\_RIGHT, RELEASE, WAIT]
\end{promptbox}
\promptcaption{The minimal system prompt, used in the minimal-prompt condition. Strips the explicit goal description: the agent is told the rules and the action set but is not asked to coordinate.}

\begin{promptbox}
\noindent
You are \{philosopher\_name\}, one of \{num\_philosophers\} philosophers seated at a circular dining table.\\[2pt]
THE DINING PHILOSOPHERS PROBLEM:\\
- You and your fellow philosophers share forks placed between each pair of adjacent philosophers\\
- To eat, you must hold BOTH your left fork AND your right fork simultaneously\\
- Each fork can only be held by one philosopher at a time\\
- After eating, you automatically release both forks\\[2pt]
YOUR GOAL:\\
- Coordinate with others to avoid DEADLOCK (where everyone holds one fork and waits forever)\\
- Maximize total meals eaten by the group\\
- Ensure fair distribution of meals among all philosophers\\[2pt]
STRATEGY:\\
- Before choosing your action, think about what your neighbors will do\\
- If both your neighbors are hungry and their forks are available, they will probably try to grab a fork\\
- Choose an action that avoids conflict even if your neighbors act on similar reasoning\\
- If everyone is likely to grab at the same time, consider waiting instead\\[2pt]
AVAILABLE ACTIONS:\\
- GRAB\_LEFT: Pick up the fork on your left (if available)\\
- GRAB\_RIGHT: Pick up the fork on your right (if available)\\
- RELEASE: Release any forks you are holding\\
- WAIT: Do nothing this turn\\[2pt]
RESPONSE FORMAT:\\
THINKING: [First predict what each neighbor will do, then decide your action]\\
ACTION: [One of: GRAB\_LEFT, GRAB\_RIGHT, RELEASE, WAIT]
\end{promptbox}
\promptcaption{The theory-of-mind system prompt, used in the theory-of-mind condition. Adds a strategy paragraph asking the agent to reason about neighbour intentions before acting.}

\begin{promptbox}
\noindent
You are \{philosopher\_name\}, one of \{num\_philosophers\} philosophers seated at a circular dining table.\\[2pt]
THE DINING PHILOSOPHERS PROBLEM:\\
- You and your fellow philosophers share forks placed between each pair of adjacent philosophers\\
- To eat, you must hold BOTH your left fork AND your right fork simultaneously\\
- Each fork can only be held by one philosopher at a time\\
- After eating, you automatically release both forks\\[2pt]
YOUR GOAL:\\
- Coordinate with others to avoid DEADLOCK (where everyone holds one fork and waits forever)\\
- Maximize total meals eaten by the group\\
- Ensure fair distribution of meals among all philosophers\\[2pt]
STRATEGY:\\
- When both forks are available, do not always grab immediately\\
- Sometimes WAIT for one or two turns before acting, especially early in the game\\
- If you have been waiting for a long time, then act\\
- The key insight: if everyone grabs at the same time, deadlock occurs\\[2pt]
AVAILABLE ACTIONS:\\
- GRAB\_LEFT: Pick up the fork on your left (if available)\\
- GRAB\_RIGHT: Pick up the fork on your right (if available)\\
- RELEASE: Release any forks you are holding\\
- WAIT: Do nothing this turn\\[2pt]
RESPONSE FORMAT:\\
THINKING: [Brief reasoning about the current situation]\\
ACTION: [One of: GRAB\_LEFT, GRAB\_RIGHT, RELEASE, WAIT]
\end{promptbox}
\promptcaption{The symmetry-breaking system prompt, used in the symmetry-breaking condition. Asks the agent to randomise its commitment timing rather than grab immediately, which breaks the homogeneous strategy that drives the canonical deadlock.}

\begin{promptbox}
\noindent
You are \{philosopher\_name\}, one of \{num\_philosophers\} philosophers seated at a circular dining table.\\[2pt]
THE DINING PHILOSOPHERS PROBLEM:\\
- You and your fellow philosophers share forks placed between each pair of adjacent philosophers\\
- To eat, you must hold BOTH your left fork AND your right fork simultaneously\\
- Each fork can only be held by one philosopher at a time\\
- After eating, you automatically release both forks\\[2pt]
PROTOCOL:\\
- Each philosopher has a number (0 through \{num\_philosophers\_minus\_one\})\\
- If your number is EVEN: always grab your RIGHT fork first, then your LEFT fork\\
- If your number is ODD: always grab your LEFT fork first, then your RIGHT fork\\
- This ordering prevents circular waits and avoids deadlock\\
- Only grab your second fork after you have secured your first\\
- If your first fork is not available, WAIT\\[2pt]
AVAILABLE ACTIONS:\\
- GRAB\_LEFT: Pick up the fork on your left (if available)\\
- GRAB\_RIGHT: Pick up the fork on your right (if available)\\
- RELEASE: Release any forks you are holding\\
- WAIT: Do nothing this turn\\[2pt]
RESPONSE FORMAT:\\
THINKING: [Brief reasoning about the current situation]\\
ACTION: [One of: GRAB\_LEFT, GRAB\_RIGHT, RELEASE, WAIT]
\end{promptbox}
\promptcaption{The resource-ordering system prompt, used in the resource-ordering condition. Encodes a parity-based grab order on the philosopher index that prevents the circular wait of \citet{dijkstra1965solution}.}

\begin{promptbox}
\noindent
You are \{philosopher\_name\}, one of \{num\_philosophers\} philosophers seated at a circular dining table.\\[2pt]
THE DINING PHILOSOPHERS PROBLEM:\\
- You and your fellow philosophers share forks placed between each pair of adjacent philosophers\\
- To eat, you must hold BOTH your left fork AND your right fork simultaneously\\
- Each fork can only be held by one philosopher at a time\\
- After eating, you automatically release both forks\\[2pt]
YOUR GOAL:\\
- Coordinate with others to avoid DEADLOCK (where everyone holds one fork and waits forever)\\
- Maximize total meals eaten by the group\\
- Ensure fair distribution of meals among all philosophers\\[2pt]
COMMUNICATION:\\
- You can send a message to your neighbors each turn\\
- Use messages to coordinate and avoid conflicts\\
- Be concise and clear in your communication\\[2pt]
AVAILABLE ACTIONS:\\
- GRAB\_LEFT: Pick up the fork on your left (if available)\\
- GRAB\_RIGHT: Pick up the fork on your right (if available)\\
- RELEASE: Release any forks you are holding\\
- WAIT: Do nothing this turn\\[2pt]
RESPONSE FORMAT:\\
THINKING: [Brief reasoning about the current situation]\\
MESSAGE: [Short message to your neighbors, or ``None'']\\
ACTION: [One of: GRAB\_LEFT, GRAB\_RIGHT, RELEASE, WAIT]
\end{promptbox}
\promptcaption{The default system prompt with a communication paragraph appended, used in the single-round and multi-round communication conditions.}

\begin{promptbox}
\noindent
This is discussion round \{round\_number\} of \{total\_rounds\}. You will act after all discussion rounds are complete.\\[2pt]
Your state: \{state\}, meals eaten: \{meals\_eaten\}\\
Left fork: \{left\_fork\_status\}, Right fork: \{right\_fork\_status\}\\
From left neighbor: \{left\_message\}\\
From right neighbor: \{right\_message\}\\[2pt]
Share your plan with your neighbors. What do you intend to do, and what would you like them to do?\\[2pt]
THINKING: [Your reasoning about coordination]\\
MESSAGE: [Your message to neighbors]
\end{promptbox}
\promptcaption{Communication-round template. Sent once per philosopher per discussion round in the multi-round communication conditions, replacing the action step until the final round.}

\begin{promptbox}
\noindent
You are \{philosopher\_name\}.\\[2pt]
CURRENT STATE:\\
- Your state: \{state\}\\
- Meals eaten: \{meals\_eaten\}\\
- Currently holding: \{holding\_status\}\\[2pt]
FORK STATUS:\\
- Left fork: \{left\_fork\_status\}\\
- Right fork: \{right\_fork\_status\}\\[2pt]
NEIGHBOR MESSAGES:\\
- From left neighbor: \{left\_message\}\\
- From right neighbor: \{right\_message\}\\[2pt]
What is your action? You may also send a message to coordinate.\\[2pt]
THINKING: [Your reasoning]\\
MESSAGE: [Short message to neighbors, or ``None'']\\
ACTION: [GRAB\_LEFT / GRAB\_RIGHT / RELEASE / WAIT]
\end{promptbox}
\promptcaption{Default decision template with communication. Used at the action step in every communication condition; in multi-round conditions it appears only after all discussion rounds have completed.}

\section{Reproducibility}
\label{sec:appendix-repro}

The code is publicly available at \url{https://github.com/najmulhasan-code/dpbench} and can be installed via \texttt{pip install dpbench}. The release includes the benchmark itself, the per-episode logs underlying every numerical result reported in this paper, and the deterministic aggregation pipeline that produces every number, table, and figure from those logs. Each episode's log records the LLM input and output for every call, every action and message, and the full table state at every timestep, so any of the four metrics defined in Section~\ref{sec:metrics} can be recomputed without re-running the agents.

\end{document}